\documentclass[sigconf,screen]{acmart}

\AtBeginDocument{%
  \providecommand\BibTeX{{%
    \normalfont B\kern-0.5em{\scshape i\kern-0.25em b}\kern-0.8em\TeX}}}

\copyrightyear{2021}
\acmYear{2021}
\setcopyright{acmcopyright}
\acmConference[HIP '21]{The 6th International Workshop on Historical Document Imaging and Processing}{September 5--6, 2021}{Lausanne, Switzerland}
\acmBooktitle{The 6th International Workshop on Historical Document Imaging and Processing (HIP '21), September 5--6, 2021, Lausanne, Switzerland}
\acmPrice{15.00}
\acmDOI{10.1145/3476887.3476905}
\acmISBN{978-1-4503-8690-6/21/09}

\acmSubmissionID{15}

\usepackage{multirow}

\definecolor{best_result}{RGB}{0,0,200}

\begin{document}

\title[Including Keyword Position in Image-based Models for Act Segmentation]{Including Keyword Position in Image-based Models for Act Segmentation of Historical Registers}


\author{Mélodie Boillet}
\orcid{0000-0002-0618-7852}
\email{boillet@teklia.com}
\affiliation{
  \institution{TEKLIA}
  \city{Paris}
  \country{France}
}
\affiliation{
  \institution{Normandie Univ, LITIS}
  \city{Rouen}
  \country{France}
}

\author{Martin Maarand}
\affiliation{
  \institution{TEKLIA}
  \city{Paris}
  \country{France}
}
\email{maarand@teklia.com}

\author{Thierry Paquet}
\orcid{0000-0002-2044-7542}
\affiliation{
  \institution{Normandie Univ, LITIS}
 \city{Rouen}
  \country{France}
}
\email{thierry.paquet@univ-rouen.fr}

\author{Christopher Kermorvant}
\orcid{0000-0002-7508-4080}
\email{kermorvant@teklia.com}
\affiliation{
  \institution{TEKLIA}
  \city{Paris}
  \country{France}
}
\affiliation{
  \institution{Normandie Univ, LITIS}
  \city{Rouen}
  \country{France}
}


\begin{abstract}

The segmentation of complex images into semantic regions has seen a growing interest these last years with the advent of Deep Learning. Until recently, most existing methods for Historical Document Analysis focused on the visual appearance of documents, ignoring the rich information that textual content can offer. However, the segmentation of complex documents into semantic regions is sometimes impossible relying only on visual features and recent models embed both visual and textual information. In this paper, we focus on the use of both visual and textual information for segmenting historical registers into structured and meaningful units such as acts. 
An act is a text recording containing valuable knowledge such as demographic information (baptism, marriage or death) or royal decisions (donation or pardon). We propose a simple pipeline to enrich document images with the position of text lines containing key-phrases and show that running a standard image-based layout analysis system on these images can lead to significant gains. Our experiments show that the detection of acts increases from 38 \% of mAP to 74 \% when adding textual information, in real use-case conditions where text lines positions and content are extracted with an automatic recognition system.

\end{abstract}



\begin{teaserfigure}
  \centering
  \includegraphics[width=0.9\linewidth]{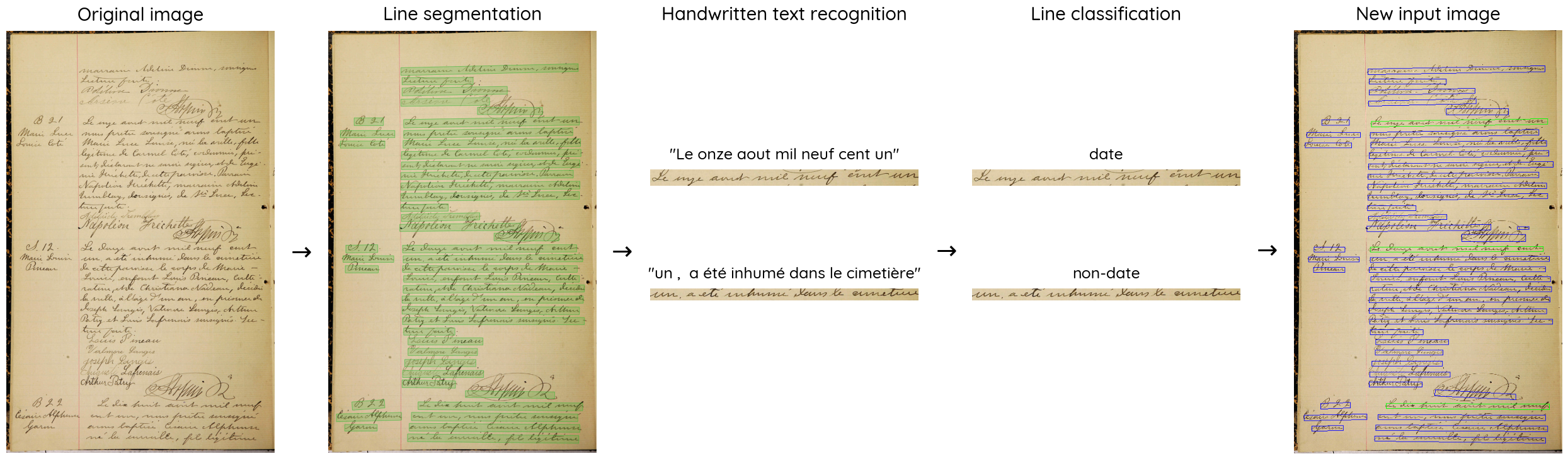}
  \caption{Pipeline used to enrich a document image with the lines positions of key-phrases. The input image is first segmented into lines and transcribed by a Handwritten Text Recognition system. Text lines positions are added to the input image, those containing key-phrases in a different color. Act segmentation is then performed by a standard semantic image segmentation model using Deep Neural Networks.}
  \label{fig:pipeline}
\end{teaserfigure}

\maketitle

\section{Introduction}
\label{sec:introduction}
With the major improvement of deep learning methods, the analysis of document images has become more and more effective. This amelioration allows the document processing community to open up to documents with more complex layouts such as historical manuscripts. Even if additional challenges emerge with historical documents such as a low quality and degradation of the documents or a bad digitization process, current proposed systems have shown high performances for various tasks such as baseline detection \cite{Gruning2018}, line segmentation \cite{dhSegment} \cite{Boillet2020} \cite{Mechi2019} and more recently act segmentation \cite{Prieto2020} \cite{Tarride2019}.


Registers are very common types of historical documents which contain lists of records, called acts, referring to persons, objects or events. They can be presented in tabular format or in the form of a sequence of texts. In the case of royal registers, medieval cartularies or parish and civil registers, the records are textual segments composed of one or more paragraphs. To deal with the act segmentation problem, most existing methods use either the textual content of the documents or the visual one. Recent systems based on heuristic rules or neural networks rely only on visual features of the images to detect the acts, ignoring the text of the documents \cite{Tarride2019}. However, historical documents can have rich textual content which can allow a better detection process. Inspired by the idea proposed in \cite{Yang2017} to incorporate the detected texts during the layout segmentation of modern documents, our work focuses on the use of both modalities at the input of a deep learning segmentation system to improve the act segmentation of historical documents.

This paper is organized as follows: Section \ref{sec:related_work} reviews existing systems dealing with act segmentation in historical documents as well as systems using multiple modalities; Section \ref{sec:datasets} presents the datasets and the document recognition system used in our experiments; Section \ref{sec:proposal} explains the method we propose and experimental evaluations are presented in Section \ref{sec:results}.

\section{Related work}
\label{sec:related_work}
Yang \textit{et al.} \cite{Yang2017} were amongst the first to propose a multi-modal Fully Convolutional Network for extracting semantic structures from modern documents. To help distinguish between similar classes like paragraphs and lists, they incorporate textual information using a text embedding map concatenated before the last convolution of the model. Adding this map did not show a significant improvement in real use-case conditions. Following this idea, Barman \textit{et al.} \cite{Barman2020} proposed a system capable of finely segmenting historical newspapers and  handling layout variation across time. They use the same textual representation as in \cite{Yang2017} but on tokens produced by an OCR process instead of sentences which is more realistic. They showed that adding the text embedding maps at the beginning of the network yields better performances. \\

In \cite{Tarride2019}, Tarride \textit{et al.} combined rules and a neural network to segment French parish registers into acts. They first detect the signatures of the priests located at the end of each act using a U-shaped neural network (dhSegment \cite{dhSegment} or \{LARU\}-Net \cite{Gruning2018}) before using a rule-based system to generate the acts. Even if they obtained a system with 80 \% of recall at act level, their method mainly relies on the hypothesis that each act ends with a signature, which is not always the case, and if present, the signature may not always be detected by the automatic system. The method we propose also includes rule-based features but combined with the original image.

Prieto \textit{et al.} \cite{Prieto2020} also studied the case where the graphical appearance of the images is not sufficient to segment medieval charters into acts. Unlike \cite{Tarride2019}, they not only aim at detecting the acts, but also seek to classify them as beginning, center, or end of act, or full act. They use Probabilistic Index Map to build additional features based on the textual content, then the graphical and textual features are merged to have a single input for the segmentation system. They show that adding textual content can help the act segmentation and that adding prior knowledge helps to further improve the performances (73 \% to 88 \% of F-value). Our method is based on the same basic idea however we use a simpler method to build the textual features. \\

The act segmentation problem can also be defined as a page stream segmentation problem: each act starting on a page can continue and end on the following one. Some works in this domain have investigated the combination of textual and visual features to classify the documents pages. In \cite{Wiedemann2018}, they use a combination of two Convolutional Neural Networks (CNN), one based on text data and one based on image scans to classify the pages. The parameters are then combined and fed to a Multi-Layer Perceptron for the final classification. This combination  increased the performances compared to a single CNN based either on text or image.

\section{Datasets and Transcription system}
\label{sec:datasets}
\begin{table}[t]
    \caption{Number of pages, transcribed lines and acts per type in the Balsac and Himanis datasets}
    \label{tab:datasets_stats}
    \setlength\tabcolsep{4.5pt}
    \begin{tabular}{ll|rrrrrr}
        \toprule
        \multicolumn{2}{c|}{\multirow{2}{*}{\textsc{Dataset}}} & \multirow{2}{*}{\textsc{Pages}} & \multirow{2}{*}{\textsc{Lines}} & \multicolumn{4}{c}{\textsc{Acts}}\\
        & & & & \textit{full} & \textit{start} & \textit{center} & \textit{end} \\
        \midrule
        \multirow{3}{*}{\textsc{Balsac}} & train & 730 & 36,941 & 1,474 & 503 & 2 & 487 \\
        & dev & 90 & 4,592 & 181 & 66 & 1 & 58 \\
        & test & 91 & 4,323 & 173 & 62 & 0 & 52 \\
        \midrule
        \multirow{2}{*}{\textsc{Himanis}} & train & 132 & - & 144 & 46 & 21 & 40 \\
        \multirow{2}{*}{\textsc{- Act}} & dev & 19 & - & 29 & 3 & 3 & 2 \\
        & test & 411 & - & 172 & 203 & 115 & 196 \\
        \midrule
        \multirow{2}{*}{\textsc{Himanis}} & train & - & 18,504 & - & - & - & - \\
        \multirow{2}{*}{\textsc{- GMV}} & dev & - & 2,367 & - & - & - & - \\
        & test & - & 2,241 & - & - & - & - \\
        \bottomrule
    \end{tabular}
\end{table}

For our experiments, we used two datasets, Balsac \cite{Vezina2020} and Himanis-Act \cite{bluche2017} and we trained a handwritten document recognition system for each of them. More details on these datasets and on the document transcription systems are given in the following paragraphs.

\subsection{Datasets}
\paragraph{Balsac Dataset} 

For the last 50 years, the BALSAC project\footnote{https://balsac.uqac.ca/} has been building and consolidating a major database on the Quebec population, covering over four centuries and five million individuals. Over the course of the project, it became increasingly apparent that manual or semi-automatic operations were no longer efficient enough to digitize, integrate and link millions of records. Fortunately, advances in handwritten text recognition and named entity extraction open up promising new avenues for automatic transcription and information extraction. At the current stage of the project, the birth, marriage and  death records for the Quebec population between 1850 and 1916  must  be integrated into the BALSAC database. They represent 1,995,646 of digitized pages from 1,985 different parishes, in a total of 44,742 registers. Each type of act must be precisely located in the register in order to extract the corresponding information (date, person names, location, occupation, role).

In order to train automatic document processing models, a sample of the corpus has been  annotated. The annotated Balsac dataset consists in 911 images (single or double pages) with act annotations and line level transcriptions. To handle acts spread over multiple pages the acts are annotated with four classes: \textit{full act}, \textit{act start}, \textit{act center} and \textit{act end}. The statistics of this dataset are presented in Table \ref{tab:datasets_stats}.

\paragraph{Himanis-Act Dataset}

The Himanis dataset is extracted from the \textit{Chancery} corpus, a large collection of registers produced by the French royal chancery. It is composed of 80,000 images containing charters given by the King of France (14th-15th c.). These documents record royal decisions like donations or pardons and are organized into acts. 

The annotated Himanis-Act dataset \cite{Prieto2020} consists in a sample of 739 pages from the Himanis dataset, annotated at act level. The size of the acts is highly variable and some acts can span several pages, therefore the same four classes as for the Balsac dataset have been used. To run our experiments, we used the split proposed in \cite{Prieto2020} obtained after eliminating pages containing no information such as blank pages so as to be comparable to their results. This dataset is only used to train and evaluate the act segmentation system since the text line segmentation and transcription annotations are not available. The final split is also presented in Table \ref{tab:datasets_stats}.

The annotated Himanis-GMV dataset is composed of 1,435 pages extracted from the Himanis dataset, automatically aligned at line level from printed editions \cite{guerin1881, viard1899, morchene2005} with Transkribus \cite{transkribus}. After the alignment process, 23,112 text lines are  available for training and evaluating the handwriting recognition system, the different split sizes are given on Table \ref{tab:datasets_stats}. This dataset is only used for training the HTR system since the act segmentation is not available. 

\subsection{Handwritten document transcription systems}
We describe in this section the text line segmentation  and the handwriting recognition systems used during the automatic act segmentation experiments. \\

\subsubsection{Line segmentation}
\label{sec:line-segmentation}

The first step to process the images consists in segmenting the pages into text lines. For this task, we used Doc-UFCN, the image segmentation model described in \cite{Boillet2020}. In order to create a generic model that can be used on different datasets, the model was trained on 9 historical datasets including Balsac but not Himanis since it contains no annotations at line level. Images were resized such that their largest side was 768 pixels, keeping the original aspect ratio and the original annotations were normalized. Additional training details can be found in \cite{Boillet2021}. Another text line segmentation system based on dhSegment \cite{dhSegment} was also trained to provide a reference comparison with the proposed system.

\begin{table}[t]
    \centering
    \caption{Line segmentation evaluations on Balsac. Doc-UFCN shows significantly higher mAP values that indicate a better detection of lines}
    \label{tab:line_segmentation}
    \setlength\tabcolsep{6pt}
    \begin{tabular}{l|rrrr}
        \toprule
        \textsc{System} & \textsc{IoU} & \textsc{AP50} & \textsc{AP75} & \textsc{mAP} \\
        \midrule
        \textsc{Doc-UFCN} & \color{best_result}0.87 & \color{best_result}0.98 & \color{best_result}0.91 & \color{best_result}0.76 \\
        \textsc{dhSegment} & 0.74 & 0.94 & 0.54 & 0.51 \\
        \bottomrule
    \end{tabular}
\end{table}

\sloppy
Both line segmentation models were evaluated using the Intersection-over-Union metric, the standard pixel level metric for segmentation tasks. The models were also evaluated using the Average Precision metric that quantifies the amount of correctly detected objects whereas Intersection-over-Union considers only the amount of correctly predicted pixels. To compute the average precision, the predicted and ground-truth objects are first paired according to their IoU scores such that only one predicted object can be paired with a ground-truth one and conversely. The Precision-Recall curve is then computed, interpolated and the Average Precision (AP) is defined as the area under this curve. We present the AP results obtained for two values of IoU threshold: 50 \% (AP50) and 75 \% (AP75). In addition, the AP averaged over IoU values in $[$0.5;0.95$]$ is also computed and presented as mAP. Results are presented on Table \ref{tab:line_segmentation} for the Balsac dataset only, since no manual annotation is available for the Himanis datasets. The results show that Doc-UFCN outperforms dhSegment for all metrics.

\subsubsection{Handwritten text recognition}
\label{sec:handwritten-text-recognition}

\fussy
The Handwritten Text Recognition (HTR) is applied on the detected text lines and outputs the corresponding text. The HTR engine is built with the Kaldi toolkit\footnote{https://kaldi-asr.org/}, which is based on a DNN-HMM (Deep Neural Network - Hidden Markov Model) model. Our model is similar to the one described in \cite{arora2019using}.
We trained a Kaldi model on Balsac dataset following the split presented in Table \ref{tab:datasets_stats}, no extra data is used for the language model nor the optical model. For Himanis, we trained a model on the Himanis-GMV annotated dataset presented above. For both models the input lines are resized to a height of 40 pixels while keeping the aspect ratio. In addition, lines with similar widths are grouped together for more efficient training. For comparison, a HTR+ model was trained using the standard Transkribus service \cite{transkribus}.

The performances of the HTR models are described in Table \ref{tab:htr_results}. Regarding our HTR system (Kaldi), the results obtained for both datasets are similar. Even if the models show relatively high Word Error Rates (WER), we believe they are able to predict a good enough transcription for keyword detection and act segmentation. The Transkribus training interface  provides an evaluation for CER only and can not be directly compared to ours since the train/dev/test splits are not identical but results are in the same range. Transkribus model shows higher CER on the training set due to the data augmentation process.

\begin{table}[t]
    \centering
    \caption{Handwritten Text Recognition evaluations on different sets of the Balsac and Himanis-GMV annotated datasets. Results of a HTR+ system trained with Transkribus are given for reference but not directly comparable with our system since the train/dev splits are not identical}
    \label{tab:htr_results}
    \setlength\tabcolsep{6pt}
    \begin{tabular}{ll|rr|rr}
        \toprule
        \multicolumn{2}{c|}{\multirow{2}{*}{\textsc{Model}}} & \multicolumn{2}{c|}{\textsc{Balsac}} & \multicolumn{2}{c}{\textsc{Himanis - GMV}} \\
        & & \textsc{CER} & \textsc{WER} & \textsc{CER} & \textsc{WER} \\
        \midrule
        \multirow{2}{*}{\textsc{Ours}} & train & 4.1 & 12.4 & 5.4 & 11.9 \\
        \multirow{2}{*}{\textsc{(Kaldi)}} & dev & 6.2 & 17.1 & 9.4 & 19.3 \\
        & test & 6.4 & 17.4 & 8.0 & 18.1 \\
        \midrule
        \multirow{2}{*}{\textsc{Transkribus}} & train$^*$ & 12.2 & NA & 9.5 & NA \\
        & dev$^*$ & 9.5 & NA & 5.3 & NA \\
        \bottomrule
    \end{tabular}
\end{table}

\section{Problem analysis and act segmentation method}
\label{sec:proposal}
\subsection{Possible approaches}
In this section, we analyse the different possible approaches for act segmentation, based on text and layout.

\paragraph{Keyphrase-based textual segmentation}

The key-phrase-based approach segments the text based only on the automatic transcription, using hand-crafted rules such as regular expressions. Depending on the regularity of document layout and on its content, this approach could yield  good enough results. For example, in the Balsac dataset, most of the acts start with a date, such as \texttt{"Le trente un janvier,  mil neuf"} and usually there are no other dates mentioned in the acts. Preliminary experiments with this method showed that, in the case of single page images with simple layout, up to 90 \% of the act type prediction could be correct. However, only 60 \% of the pages containing acts were detected because of HTR errors. Adapting the rules to all the possible HTR errors would have been too time-consuming so we did not pursue in this direction. Regarding the Himanis corpus, we could not apply a text-based extraction method since the acts do not start with a date nor the same words. 

\paragraph{Raw image-based segmentation}
\label{sec:proposal_raw_images}
The raw image-based approach considers only the image of the page to detect and type the acts. Using deep neural networks for semantic image segmentation allows to train an act segmentation system directly on raw images, without the burden of manually defining rules as in the text-based approach. Preliminary experiments, reported in Section \ref{sec:results}, showed that despite good results on non-consecutive classes (\textit{act start} or \textit{act end} objects), too many consecutive full acts were merged, showing the limit of a purely image-based approach. \\

From our preliminary experiments, the main limitations of previous methods can be summarized as follows: (1) Heuristic text-based rules do not work for pages with a complex layout; (2) Heuristic text-based rules are not applicable when the context is variable; (3) Visual-based method struggles to split consecutive full acts whose visual aspects are similar.

\subsection{Proposed method for act segmentation}
We propose a strategy to address this challenges and to have a system capable of detecting and typing the acts by jointly using visual information and textual content of the pages. It consists in using both previous methods together by using the basic idea of the heuristic method as an additional input to the deep learning system. Even if detecting key-phrases from automatic transcription may not be sufficient by itself, we believe that adding textual content to the input image, even if not perfect, can help the network to better detect the acts. Our proposal to have a multi-modal input consists in the following steps as illustrated on title page Figure \ref{fig:pipeline}:
\sloppy
\begin{enumerate}
    \item Text line segmentation of the images;
    \item Handwritten text recognition on the detected lines;
    \item Classification of the lines according to their content: whether they contain a key-phrase or not;
    \item Line drawing on the original input images: green for a key-phrase and blue for the others;
    \item Using the image augmented with key-phrase line positions as an input for a deep neural network for semantic image segmentation.
\end{enumerate}

\fussy
In the rest of this section, we give more details on the line classification and act segmentation steps.

\paragraph{Line classification}

The line classification is performed using heuristic rules. The model uses the predicted transcriptions as input and predicts if a text line is the first line of an act based on the presence of key-phrases defined manually.

For the Balsac dataset, most of the acts start with a date, therefore the rule is to count the number of words that are numbers or months for the line to be considered as a date. Experimentally, three words seemed to be enough for the line to be considered as containing a date. 

The task for Himanis acts is more complex as they do not always start with the same words. Therefore, we looked at the first lines of the training acts and kept the most frequent key-phrases (for example \texttt{"dei gratia francorum rex"} or \texttt{"par la grace de dieu roys de france"}). If a key-phrase is included in a line, it is considered as being at the beginning of an act.

\paragraph{Act segmentation}

Lastly, to detect the acts from enriched images, we used the same model as for the line segmentation step but with the act positions and classes as targets. Lines containing key-phrases can be included in the image segmentation model in different ways: 
\begin{enumerate}
    \item by drawing only key-phrase lines on input image;
    \item by drawing the lines into a 4th input channel, therefore combining the original image and the layout;
    \item by drawing all lines on input image with a different color for date/key-phrase lines.
\end{enumerate}

For the training of the segmentation model, the line polygons and transcriptions used as input do not come from the annotations but from the predictions of the models described in previous sections. This way the training and evaluation conditions are more similar regarding the uncertainty on the line position and on the textual content.

\section{Experimental evaluation}
\label{sec:results}
In this section, we present an experimental evaluation of the different methods for act segmentation on the two datasets.

\begin{table}[t]
    \centering
    \caption{First line classification evaluations in Precision, Recall and F1-score (\%) for Balsac and Himanis-Act}
    \label{tab:classification_results}
    \setlength\tabcolsep{6pt}
    \begin{tabular}{l|rrr|rrr}
        \toprule
        \multirow{2}{*}{\textsc{Split}} & \multicolumn{3}{c|}{\textsc{Balsac}} & \multicolumn{3}{c}{\textsc{Himanis - Act}} \\
        & \textsc{P} & \textsc{R} & \textsc{F1} & \textsc{P} & \textsc{R} & \textsc{F1}  \\
        \midrule
        train & 69.2 & 87.0 & 77.1 & 78.6 & 64.7 & 71.0 \\
        dev   & 71.7 & 87.0 & 78.6 & 81.0 & 53.1 & 64.2 \\
        test  & 68.4 & 85.9 & 76.1 & 67.6 & 85.5 & 75.5 \\
        \bottomrule
    \end{tabular}
\end{table}

\subsection{Line classification}
Here, we evaluate the line classification based on the key-phrase extraction. Table \ref{tab:classification_results} shows the precision, recall and F1-score of class "first line". Only class "first line" is given since it is the only class bringing information to the act segmentation and the class distribution is highly unbalanced. For the Balsac dataset, the results are stable between the three sets and the recall is high which is favorable for including this information into the visual system. For Himanis-Act, the recall is lower and the results vary across the sets which implies that the task is more complex and that the key-phrase detection is less reliable.

\subsection{Act segmentation}
\label{sec:results_act_segmentation}

\begin{table}[t]
    \setlength\tabcolsep{4.5pt}
    \centering
    \caption{Evaluations of three act segmentation models (RawImage, Image+TextMask and Image+KeyLines) on the Balsac and Himanis-Act datasets according to a pixel metric (IoU) and three object detection metrics (AP50, AP75 and mAP)}
    \label{tab:act_segmentation_results}
    \begin{tabular}{lll|rrrr}
        \toprule
        \textsc{Dataset} & \textsc{Model} & \textsc{Class} & \textsc{IoU} & \textsc{AP50} & \textsc{AP75} & \textsc{mAP} \\
        \midrule
         \multirow{13.1}{*}{\textsc{Balsac}} & \multirow{3}{0.1\textwidth}{\textsc{Doc-UFCN Raw images}} & \textit{full} & \color{best_result}0.84 & 0.57 & 0.37 & 0.38 \\
        &  & \textit{start} & 0.58 & 0.86 & 0.85 & 0.76 \\
        & & \textit{end} & 0.58 & 0.85 & 0.64 & 0.59 \\
        \cmidrule{2-7}
        & \multirow{3}{0.1\textwidth}{\textsc{Doc-UFCN  Image + TextMask}} & \textit{full} & 0.83 & 0.44 & 0.27 & 0.27 \\
        & & \textit{start} & \color{best_result}0.64 & 0.90 & 0.83 & 0.75 \\
        & & \textit{end} & \color{best_result}0.61 & \color{best_result}0.89 & 0.70 & \color{best_result}0.65 \\
        \cmidrule{2-7}
        & \multirow{3}{0.1\textwidth}{\textsc{Doc-UFCN Image + KeyLines}} & \textit{full} & 0.82 & \color{best_result}0.89 & \color{best_result}0.81 & \color{best_result}0.74 \\
        &  & \textit{start} & 0.58 & \color{best_result}0.90 & \color{best_result}0.87 & \color{best_result}0.78 \\
        & & \textit{end} & 0.54 & 0.86 & \color{best_result}0.73 & 0.63 \\
        \cmidrule{2-7}
        &\multirow{3}{0.1\textwidth}{\textsc{dhSegment Image + KeyLines}} & \textit{full} & 0.75 & 0.19 & 0.08 & 0.08 \\
        & & \textit{start} & 0.48 & 0.81 & 0.49 & 0.47 \\
        & & \textit{end} & 0.41 & 0.80 & 0.48 & 0.48 \\
        \midrule

        \multirow{7.4}{*}{\textsc{Himanis}} & \multirow{4}{0.1\textwidth}{\textsc{Doc-UFCN Raw Images}}  & \textit{full} & 0.61 & \color{best_result}0.75 & \color{best_result}0.73 & \color{best_result}0.70 \\
        \multirow{7.4}{*}{\textsc{- Act}} & & \textit{start} & \color{best_result}0.76 & \color{best_result}0.84 & \color{best_result}0.82 & \color{best_result}0.77 \\
        & & \textit{center} & \color{best_result}0.88 & \color{best_result}0.84 & \color{best_result}0.83 & \color{best_result}0.83 \\
        & & \textit{end} & \color{best_result}0.73 & \color{best_result}0.73 & \color{best_result}0.65 & \color{best_result}0.62 \\
        \cmidrule{2-7}
        & \multirow{4}{0.1\textwidth}{\textsc{Doc-UFCN Image + KeyLines}} & \textit{full} & \color{best_result}0.64 & 0.54 & 0.51 & 0.49 \\
        & & \textit{start} & 0.68 & 0.69 & 0.64 & 0.60 \\
        & & \textit{center} & 0.84 & 0.80 & 0.80 & 0.80 \\
        & & \textit{end} & 0.70 & 0.64 & 0.63 & 0.58 \\
        \bottomrule
    \end{tabular}
\end{table}

Table \ref{tab:act_segmentation_results} presents the results of different act segmentation systems: Doc-UFCN trained on \textit{Raw Images} (Section \ref{sec:proposal_raw_images}), Doc-UFCN trained with 4 input channels containing the raw image and the polygons mask (\textit{Image+TextMask}) and finally Doc-UFCN and dhSegment \cite{dhSegment} trained on the raw images with the polygons of the text lines drawn using two colors depending on the presence of a key-phrase (\textit{Image+KeyLines}). For Himanis-Act, we only present the results of the visual (\textit{RawImages}) and \textit{Image+KeyLines} models.

\subsubsection{Balsac}

For the Balsac dataset, we do not report the results of the \textit{act center} class since there are none in the test set. According to Table \ref{tab:act_segmentation_results}, the results are on average better for the \textit{Image+KeyLine} model. For the \textit{act start} and \textit{act end} classes, the three systems are almost equivalent for all metrics. On the contrary, we see that adding the date information directly to the input image did improve the performance on the \textit{full act} class by 36 percentage points. This leads to a better separation of consecutive full acts in the predictions which was one of our main challenges. \\

\sloppy
For comparison, we present the results of dhSegment using the \textit{Image+KeyLines} enriched input images trained in the same conditions as Doc-UFCN. According to Table \ref{tab:act_segmentation_results}, dhSegment does not segment the full acts very well. Indeed, it correctly finds them as reported by the high IoU value but fails to split consecutive ones since it shows only 8.08 \% of mAP. Doc-UFCN clearly outperforms it, especially on the \textit{full act} class where it shows a mAP value of 73.52 \%. 

\begin{table}[t]
    \centering
    \caption{Results of Doc-UFCN and Prieto \textit{et al.} \cite{Prieto2020} models obtained on Himanis-Act dataset with and without the textual information}
    \label{tab:himanis_results}
    \setlength\tabcolsep{6pt}
    \begin{tabular}{ll|rr}
        \toprule
        \multicolumn{2}{c|}{\multirow{2}{*}{\textsc{Model}}} & \multirow{2}{*}{\textsc{Image}} & \textsc{Image +} \\
        & & & \textsc{Text} \\
        \midrule
        \multirow{3}{*}{\textsc{Doc-UFCN}} 
        & train & 0.96 & 0.96 \\
        & dev & 0.96 & 0.91 \\
        & test & \color{best_result}0.90 & 0.88 \\
        \midrule
        \textsc{Prieto et al. \cite{Prieto2020}} & test & 0.80 & 0.88\\
        \bottomrule
    \end{tabular}
\end{table}

\subsubsection{Himanis}

\fussy
For this dataset, the \textit{RawImages} results are clearly better than the \textit{Image+KeyLines} ones. We believe that this is due to the following reasons. First the Himanis-Act textual content is more complex and diverse than the Balsac dataset content which is very standardized. Moreover,  we could find many nested acts in the dataset (\textit{Vidimus}), which can add confusion to the system. Defining the key-phrases was more complex and it can be seen from Table \ref{tab:classification_results}, that recall is low even for the train set, leading to unreliable textual features for training the segmentation model. In addition, both the line segmentation and HTR models were not trained on the Himanis-Act dataset. The line segmentation model was trained on similar data but without any images for the Himanis-Act dataset. The same applies for the HTR system that was not directly trained on the same volumes, which can create a mismatch between training and testing conditions. \\

In addition to these experiments, we compared our results with the state-of-the-art ones of \cite{Prieto2020}. They tested different configurations with and without the textual content to detect where the acts end. To be comparable, we used the same evaluation scheme. The evaluation is done using the Transkribus Baseline Evaluation Scheme (TBES) \cite{Gruning2017}. This tool is used to evaluate the baseline detection, therefore to use it they define the baseline of the acts as the horizontal straight line at the end of a \textit{full act} or an \textit{act end} predicted object. To be in accordance with their results, we used the same \textit{tolerance} value of 128 pixels.

From Table \ref{tab:himanis_results}, we can see that Doc-UFCN  using only the image outperforms the state-of-the-art results. Indeed, when compared to the visual system of \cite{Prieto2020}, our method outperforms it by up to 10 percentage points. In addition, we see that both models using the textual content behave alike and are worse than our system based only on visual information. For this dataset, it seems to be better to focus on visual features with a robust deep learning system than trying to add textual content that is too unreliable.

\section{Conclusion}
In this paper, we presented a simple pipeline used to enrich input images with the textual content of documents. These enhanced images allow to perform a semantic act segmentation task using both visual and the position of text lines containing manually defined key-phrases. We showed that using these images can  improve the detection of acts, especially consecutive ones. On Balsac dataset, for which key-phrases detection rules could be reliably defined, using such images increases the detection of acts from 38 \% of mAP to 74 \% compared to a standard baseline model.
As a future work, we will compare this approach to pre-trained models recently proposed for taking into account both the textual content and its layout \cite{layoutlm}.

\begin{acks}

This work is part of the \textit{HOME History of Medieval Europe} research project, supported by the European JPI Cultural Heritage and Global Change (Grant agreements No. ANR-17-JPCH-0006 for France, MSMT-2647\/2018-2/2, id. 8F18002 for Czech Republic and PEICTI Ref. PCI2018-093122 for Spain). It is also supported by the i-BALSAC project with the Université du Québec in Chicoutimi. Mélodie Boillet is partly funded by the CIFRE ANRT grant No. 2020/0390.

\end{acks}

\bibliographystyle{ACM-Reference-Format}
\bibliography{acmart}

\end{document}